\documentclass[10pt,journal,compsoc]{IEEEtran}

\usepackage{bm}
\usepackage{times}
\usepackage{epsfig}
\usepackage{diagbox}
\usepackage{graphicx}
\usepackage{amsmath}
\usepackage{mathrsfs}
\usepackage{amssymb}
\usepackage{threeparttable}
\usepackage{adjustbox}
\usepackage{textcomp}
\usepackage{multirow}
\usepackage{rotating}
\usepackage{makecell}
\usepackage{graphics} 
\usepackage{enumitem}
\usepackage{setspace}
\usepackage[colorlinks]{hyperref}

\ifCLASSOPTIONcompsoc
  \usepackage[nocompress]{cite}
\else
  \usepackage{cite}
\fi
\usepackage{array}

\usepackage{url}

\begin{document}
%

\title{SchiNet: Automatic Estimation of Symptoms of Schizophrenia from Facial Behaviour Analysis}

%
%


\author{Mina~Bishay,
        Petar~Palasek,
        Stefan~Priebe,
        and~Ioannis~Patras  
\IEEEcompsocitemizethanks{\IEEEcompsocthanksitem Mina Bishay and Ioannis Patras are with the School of Electronic Engineering and Computer Science, Queen Mary University of London, UK. \protect\\
E-mail:\{m.a.t.bishay,i.patras\}@qmul.ac.uk.
\IEEEcompsocthanksitem Petar Palasek was with the School of Electronic Engineering and Computer Science, Queen Mary University of London, UK. He is currently working as a research scientist at MindVisionLabs LTD. \protect\\ 
E-mail:p.palasek@qmul.ac.uk.
\IEEEcompsocthanksitem Stefan Priebe is with the Unit for Social and Community Psychiatry (WHO Collaborating Centre for Mental Health Service Development), Newham Centre for Mental Health, Queen Mary University of London, London, UK. \protect\\ 
E-mail:s.priebe@qmul.ac.uk.}

}

\IEEEtitleabstractindextext{%
\begin{abstract}


Patients with schizophrenia often display impairments in the expression of emotion and speech and those are observed in their facial behaviour. Automatic analysis of patients' facial expressions that is aimed at estimating symptoms of schizophrenia has received attention recently. However, the datasets that are typically used for training and evaluating the developed methods, contain only a small number of patients (4-34) and are recorded while the subjects were performing controlled tasks such as listening to life vignettes, or answering emotional questions. In this paper, we use videos of professional-patient interviews, in which symptoms were assessed in a standardised way as they should/may be assessed in practice, and which were recorded in realistic conditions (i.e. varying illumination levels and camera viewpoints) at the patients' homes or at mental health services. We automatically analyse the facial behaviour of 91 out-patients -- this is almost 3 times the number of patients in other studies -- and propose SchiNet, a novel neural network architecture that estimates expression-related symptoms in two different assessment interviews. We evaluate the proposed SchiNet for patient-independent prediction of symptoms of schizophrenia. Experimental results show that some automatically detected facial expressions are significantly correlated to symptoms of schizophrenia, and that the proposed network for estimating symptom severity delivers promising results.

\end{abstract}


}

\maketitle

\IEEEdisplaynontitleabstractindextext

%
\IEEEpeerreviewmaketitle

\ifCLASSOPTIONcompsoc
\IEEEraisesectionheading{\section{Introduction}\label{sec:introduction}}
\else
\section{Introduction}
\label{sec:introduction}
\fi


\IEEEPARstart{S}{chizophrenia} is a severe mental illness affecting not only the patients, but also their families and the society as a whole. Patients with schizophrenia often show impairment in the expression of emotion and speech in comparison to non-patients \cite{tremeau2006review}  -- this is manifested in their facial expression \cite{mandal1998facial}, vocal expression \cite{leentjens1998disturbances, murphy1990prosodic}, and expressive gestures \cite{brune2008nonverbal, troisi1998non}. Patients can also show impairment in the non-verbal behaviour that invites social interaction during clinical and nonclinical interviews \cite{lavelle2014nonverbal}. Non-verbal behaviour was found to change during interviews according to symptom severity \cite{dimic2010non, lavelle2012nonverbal, worswick2017negative}. For instance, patients with high symptom severity tend to avoid interaction by nodding less, smiling less, and looking less at the interviewer \cite{dimic2010non}. Such impairments present valuable information for the psychiatrists, as they can be used for diagnosis and symptom assessment. However, behaviour analysis is time-consuming in research settings and subjective in clinical settings -- this calls for the development of automatic analysis tools. 



Recently, there has been a growing interest in studying behaviour differences in groups of patients with schizophrenia and healthy controls, as well as diagnosing schizophrenia using Automatic Facial Expression Analysis (AFEA) \cite{alvino2007computerized, tron2015automated, wang2008automated, tron2016facial}. The reason for the interest is that AFEA allows objective and fast measurement of facial expressions and that can be valuable for both research and diagnosis. However, the datasets that are used in current works contain only a few patients (4-34 patients) and are recorded while they were performing controlled tasks, such as listening to life vignettes, or answering emotional questions. In addition, the tools that are typically used/proposed for AFEA perform well primarily in a specific, controlled environment that is hard to be replicated in clinics and hospitals. Furthermore, the methods proposed up to now for diagnosing schizophrenia rely on conventional hand-crafted features that have shown inferior performance in comparison to learned ones and in particular those learned by Deep Neural Networks.



In this paper, we move from controlled environments to similar to real life settings and use professional-patient interviews of symptom assessment. More specifically, we use research interviews in which symptoms were assessed in a standardised way as they should/may be assessed in real life clinical encounters. The interviews involve a selection of patients with negative symptoms -- such symptoms are particularly difficult to assess and quantify \cite{savill2015negative}. The interviews were recorded either at the patients' homes or at the premises of mental health services across the UK. The collected videos have a wide range of camera viewpoints and illumination levels that are representative of the variety of settings found in clinics. We used interviews of 91 out-patients -- this is almost 3 times the highest number of patients used in other studies. 




In order to automatically analyse the videos, we propose a Deep Neural Network (DNN) architecture, called SchiNet, that analyses facial expressions and estimates symptoms of schizophrenia that are related to them. The proposed SchiNet is patient-independent and consists of two main stages. In the first stage, different DNNs are used for detecting patients' facial expressions, such as smiles and activations of facial muscles at each frame (low-level features). At the second stage, a DNN consisting of a) Gaussian Mixture Model (GMM) and Fisher Vector (FV) layers for extracting a compact statistical feature vector over the whole video interview (high-level features), and b) a regression layer is used for symptom estimation. The different sub-networks are first trained in stages and then are refined in an end-to-end fashion. 




The proposed network has been trained in a person-independent manner to predict expression-related symptoms from two commonly-used assessment interviews; Positive and Negative Syndrome Scale (PANSS) \cite{kay1987positive}, and Clinical Assessment Interview for Negative Symptoms (CAINS) \cite{horan2011development}. Experimental results show that training the Facial Expression Analysis sub-network (stage 1) ``in the wild" delivers better performance on symptom severity estimation in comparison to another state-of-the-art method \cite{bishay2017fusing} that is trained using data captured in a controlled environment. Furthermore, we show that high and statistically significant correlations between the detected expressions and the severity of several symptoms in both the PANSS and CAINS can be obtained.



The main contributions of our work are two-fold:

\begin{enumerate}[noitemsep] 
   \item We move from controlled contexts to settings that are similar to real life ones, where we analyse symptom assessment interviews of almost three times the number of patients used in previous studies.
   \item We propose a fully-automatic deep learning approach for estimating expression-related symptoms of schizophrenia in two different assessment interviews, namely PANSS and CAINS.
\end{enumerate}



The rest of the paper is organized as follows: In Section 2, we review the related literature in analysing and diagnosing schizophrenia from facial expressions. In Section 3, we introduce the clinical dataset that we used in the analysis. In Section 4, we present the proposed SchiNet for estimating symptoms of schizophrenia. Finally, in Section 5 and Section 6 we give the experimental results and the conclusions, respectively.



\section{Related Work}



Some psychiatric researches are concerned with the relation between schizophrenia and the patients' non-verbal behaviour \cite{davison1996facial, dimic2010non, lavelle2012nonverbal, troisi1998non, worswick2017negative}. To perform quantitative analysis, in these works the video intervals were manually annotated in terms of the patients' non-verbal behaviour and, subsequently, statistical analysis, such as calculation of the correlations of that behaviour with the severity of the symptoms was performed. However, manual annotation of videos is a hard and time-consuming task and requires a special training. For this reason, in the last few years there has been a growing interest in the application of Automatic Facial Expression Analysis (AFEA) methods for studying and diagnosing schizophrenia. In this section, we review related works in terms of the datasets and the AFEA methods that are used, in addition to the main objectives of these works. 




\textbf{Datasets.} Due to the difficulty and the ethical issues in the collection and management of data depicting patients' behaviour, there are only a few datasets available in the domain of schizophrenia. Two datasets are used in a number of works; the first one is collected in a mental health centre at the University of Pennsylvania (Penn), while the second at the Hebrew University of Jerusalem (HUJI). In this work we refer to the former as Penn-dataset, and the later as HUJI-dataset. The Penn-dataset consists of videos and images that are collected at two different sessions. In the first session, patients with schizophrenia and healthy controls are asked to express basic emotions at 3 different intensities. In the second session, they are recorded while listening to vignettes about a situation in their life that is presented by them before recording. Each vignette is expected to evoke 1 of 4 basic emotions; happiness, sadness, anger and fear. The number of participants in this dataset varies across different studies \cite{alvino2007computerized, wang2007quantifying, wang2008automated, hamm2011automated, hamm2014dimensional}, but it is at most 28 patients and 26 controls. The HUJI-dataset is recorded while subjects (patients and healthy controls) were participating in structured interviews. During these interviews, the participants were asked emotional questions, and also shown 20 emotional images from the International Affective Picture System. This dataset has 34 patients and 33 healthy controls, and it is used in \cite{tron2015automated, tron2016facial, tron2016differentiating}. 




\textbf{AFEA methods.} Different methods have been used/proposed in the literature for analysing patients' facial behaviour. In \cite{alvino2007computerized}, Alvino {\em et al.} detected static emotional expressions by measuring a deformation between a neutral face and a face with expression, which was then classified using an SVM classifier. In \cite{wang2008automated}, Wang {\em et al.} proposed the use of temporal facial information (as opposed to only static) for analysing emotional expressions. To do so, first an SVM classifier trained using geometric features was applied for estimating the probabilities of expressions at each video frame and then a sequential Bayesian estimation, with the goal of propagating probabilities throughout the video, was applied. In \cite{hamm2011automated, hamm2014dimensional}, Hamm {\em et al.} moved from analysing basic emotions to detection of 15 Action Units (AUs) at every frame of the sequence. The AUs were detected by training a Gentle Adaboost classifier using geometric and texture features. A problem with those AFEA methods is that they were trained on frontal views and on evoked expressions from professional actors. As we will show, such methods are not suitable for analysis of non-verbal behaviour in uncontrolled conditions, such as professional-patient interviews, as they are not robust to variabilities in recording factors such as camera viewpoint and illumination levels. Similar results are reported in other studies: For example, \cite{cao20133d} reports that the commercial 3D facial analysis tool used for detecting 23 AUs in \cite{tron2015automated, tron2016facial, tron2016differentiating}, has restrictions on the distance between the user and the camera as well as the working environment.



\textbf{Analysis.} Several studies focused on comparing a group of patients with schizophrenia to a group of healthy controls in terms of information extracted from facial expression analysis investigating the existence of differences between them. In addition, correlations between these features and flatness and inappropriateness symptoms in the SANS scale \cite{andreasen1989scale} were tested. Various features were extracted in these studies. In \cite{alvino2007computerized, wang2008automated}, the average probability of 4 emotions and neutral expression were calculated. In \cite{wang2007quantifying}, 2D geometric features and 3D curvature features were used in the comparison. In \cite{hamm2011automated}, features as frequency of some single and combined AUs were extracted, while in \cite{hamm2014dimensional} information theory measures were used as features for comparing and assessing ambiguity and distinctiveness of subjects' facial expressions. Correlations were found to be significant with the flatness symptom, and insignificant with the inappropriateness symptom. Furthermore, in \cite{tron2016differentiating} the facial activity of patients and controls, watching a set of emotionally evocative pictures, was analysed and used for differentiating flat and incongruent affects in schizophrenia. Variance analysis over the facial activity was used to measure flatness (variance in expressions) and incongruity (relative variance in response to similar stimuli).

A few studies by Tron {\em et al.} \cite{tron2015automated}\cite{tron2016facial} go beyond studying the differences in behaviour between patients and healthy controls, and more specifically, use automatic analysis of facial behaviour for diagnosis and severity estimation of some PANSS symptoms (especially flat affect). In these studies, different features were extracted and used with a two-step SVM based algorithm for the diagnosis and symptom estimation. In \cite{tron2015automated}, features related to the intensity and dynamics of each AU (e.g. frequency, activation length, change ratio) were extracted, while in \cite{tron2016facial}, clustering analysis was used over all AUs for extracting 3 flatness-related features, richness (number of facial-clusters appeared), typicality (the similarity to prototype), and cluster distribution (the activation frequency of different clusters).


We can first conclude that most of the conducted research focuses on studying behaviour differences between patients and healthy individuals and that only a couple of works address the problems of the diagnosis and symptom estimation in schizophrenia. Second, the datasets used in these works contain a relatively small number of patients and were recorded while the patients were performing controlled tasks. Third, the tools used/proposed for facial expression analysis work either on frontal views or in a specific environment. Finally, all the features used in the diagnosis and symptom estimation in schizophrenia are hand-crafted ones -- this can have implications on the performance of the regression/classification model. By contrast, we use video recordings of 91 patients in conditions that are similar to realistic symptom assessment interviews. In addition, we use statistical deep features for estimating expression-related symptoms in two different assessment interviews, PANSS and CAINS. All of the networks that we use, including the first stage that analyses facial expressions, are trained with data ``in the wild", in order for them to be robust to different recording conditions.



\begin{figure*} [t!]
\centering
\centering{\includegraphics[scale=0.44]{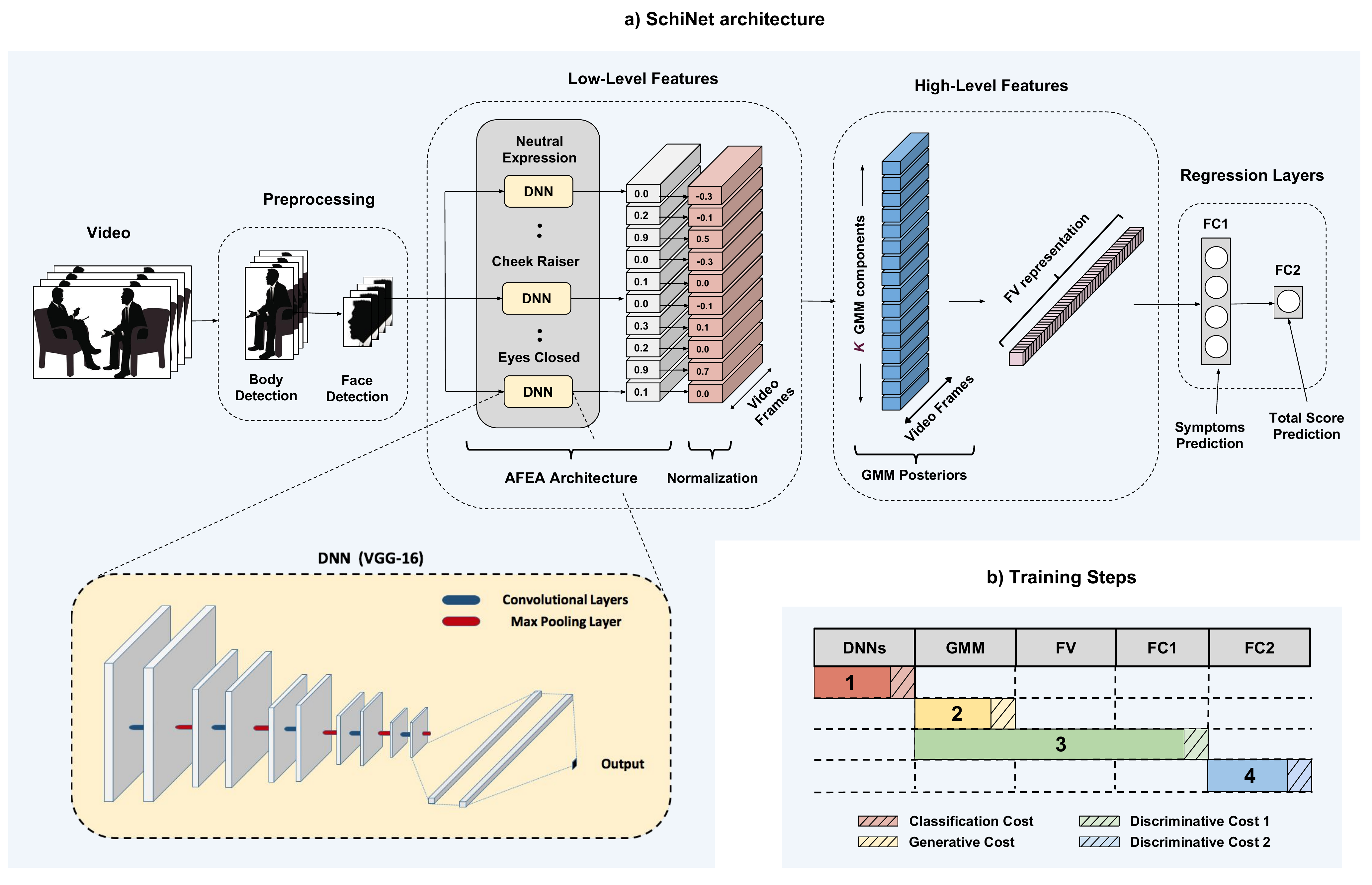}%
\label{cp}}
\hfil
\caption{ (a) The proposed SchiNet for symptom severity estimation in schizophrenia. The input is a recorded video interview of a patient during his/her symptom assessment, and the outputs are the estimated values for the expression-related symptoms and the total scale/symptoms score. Feature extraction is done over two stages, first, the video is encoded by patient facial expressions, then a compact statistical feature vector is extracted over the encoded expressions. (b) The training stages of the SchiNet.  
}
\label{fig2}
\end{figure*}


\section{Clinical Dataset of Schizophrenia}


In this work we use a dataset called ``NESS", that was collected for studying the effectiveness of group body psychotherapy on negative symptoms of schizophrenia \cite{priebe2016effectiveness}. The participants in this study were recruited from mental health services at four different places in the UK; East London, South London, Liverpool, and Manchester. In total, 275 participants were included in this study. Participants aged between 18-65, and they had a total negative symptoms score $\geq$ 18 on the PANSS interview, that is, the study focused on patients with negative symptoms. Those symptoms are typically difficult to assess and quantify \cite{savill2015negative}.  


The participants were assessed at three different stages throughout the study; BaseLine $(BL)$ -- before the start of the treatment, End of Treatment $(EoT)$ -- after completing 20 session of group body psychotherapy, and 6 Months Follow-Up $(6MFU)$ -- 6 months after the end of treatment. Each assessment interview lasted between 40 and 120 minutes, depending on the time spent by patients in speaking and recollection about the interview questions. The patients were assessed at the interview in terms of PANSS \cite{kay1987positive} including negative, positive and general psychopathology symptoms, and CAINS \cite{horan2011development} including experience-related and expression symptoms. In addition, other scales related to depression, quality of life and client satisfaction for patients with schizophrenia were also assessed. The interviews were completed in a standardised way by researchers/psychologists as they should/may be done in real life clinical encounters.


Only the assessment of the PANSS and CAINS were video-recorded from the whole interview. Most of the videos were recorded at 25 frames/s and at a resolution of 1920$\times$1080. Out of the 275 patients, 110 accepted to be recorded at BaseLine, 93 at End of Treatment, and 69 at 6 Months Follow-Up. Since the focus of this paper is building a model that estimates the symptom severity for unseen patients (i.e. a generic model), only the 110 patients recorded at the $BL$ session are used in our analysis. The average length of the recorded $BL$ interviews is 41 minutes. More information about the dataset can be found in \cite{priebe2016effectiveness}.


\section{Proposed Architecture}

\subsection{Overview}

In this section we present a deep architecture, named SchiNet, for estimating the severity of symptoms of schizophrenia from videos depicting the non-verbal behaviour of patients. Figure \ref{fig2}(a) shows an overview of the system. SchiNet takes as input a video interview for patient symptom assessment and gives as output the estimated values of expression-related symptoms and the total scale/symptoms score. Intermediate results include detection of facial expressions at frame level and statistical representations of their activations in the whole image sequence.


SchiNet performs the analysis in 4 stages; preprocessing, low-level feature extraction at frame level, high-level feature extraction at video level and symptoms regression. At the first stage, we detect the patients' faces in the video frames using a body detector \cite{liu2016ssd} and a robust face detector \cite{jang2017smilenet}. At the second stage, the face regions are cropped and passed to a bank of Deep Neural Networks (DNNs), each of which detects a certain facial expression or the activation of a certain facial Action Unit. Encoding the patients' facial behaviour at each frame is considered as the first/low-level feature extraction. At the third stage, a Gaussian Mixture Model (GMM) and a Fisher Vector (FV) layer are used to represent the patient facial behaviour over the whole video by a compact feature vector (i.e. FV representation). The FV representation is considered as the second/high-level feature extraction. Finally, the FV is fed to two fully-connected layers for estimating the symptoms and the total score.




The training of the SchiNet is done in 4 stages, as shown in Figure \ref{fig2}(b). At each stage, a different cost is optimised. In the first stage, the DNNs that detect the activation of facial expressions at each frame are trained in a supervised manner on datasets annotated with the corresponding labels. At the second stage, the network that extracts video-based representations is trained in an unsupervised manner, taking as input the sequence of the outputs of the first network when applied to the professional-patient interviews. More specifically, the distribution of the expression probabilities in a video is modelled using a GMM that is implemented as a network layer. Then, the estimated GMM parameters are used to extract a FV representation for the whole video. In the third stage the FV representations are used as input to a regression layer that estimates symptoms of schizophrenia (flat affect, poor rapport, and lack of spontaneity and flow of conversation symptoms in the case of PANSS and 4 Expression symptoms in the case of CAINS). Following \cite{palasek2017discriminative}, we refine the GMM, the FV and the first regression layer in an end-to-end fashion using a discriminative cost. Finally, in the fourth stage, we train a second regression layer that takes as input the individual symptom scores and estimates the total scale/symptoms score. The SchiNet architecture as well as the training stages are explained in detail in the following subsections.





\subsection{Preprocessing Steps}


In order to process each video in the NESS dataset, we first extract the region of interest (i.e. the patient's face) at each frame. We do so in four steps. First, we detect the patient's body at each frame using the Single Shot Detector (SSD) proposed in \cite{liu2016ssd}. We then extend the detected body-bounding box by a factor of 1.2 to ensure that the whole head is included, and then, within the resulting region, we apply SmileNet \cite{jang2017smilenet} to detecting the bounding box of the face and whether it is smiling or not. Finally, we crop and scale the detected face to a fixed resolution of 100$\times$100 for further analysis. Note that no face-registration is applied to the extracted faces prior to expression analysis.


Despite the robustness of the face detection, it still fails in some videos due to the position of the camera. In those cases, not only the face is sometimes not detected but, even in the cases that it is, it is hard to be further analysed in terms of the facial expressions. For this reason, we consider only the videos in which we can successfully detect the faces in more than $90\%$ of the frames. By doing so, we retain the videos of 91 patients out of the total 110 that participated in the baseline session.




\subsection{Low-Level Feature Extraction}
\label{low_feat}

In the second stage of the proposed method a DNN architecture is used to code the facial behaviour in terms of Facial Action Coding System (FACS) \cite{EkmanBook97}, that is, detect the activation of facial Action Units (AUs), and detect smiles, a specific facial behaviour, the absence of which is expected to be informative in the assessment of negative symptoms of schizophrenia. FACS has been extensively used for facial expression analysis in different contexts \cite{valstar2006spontaneous, williamson2014vocal, cohn2014automated, martinez2017automatic}. 




\textbf{Proposed AFEA Method.} Detection of facial AUs, i.e. detection of the activation of certain facial muscles, is recently being treated as a pattern recognition problem, where one trains in a supervised manner classifiers that receive as input an image, or features extracted from it, and give at the output a set of binary labels, as many as the AUs that the method detects. In recent years the low-level feature extraction and the classifiers are replaced by Convolutional Neural Networks (CNNs) \cite{bishay2017fusing, jaiswal2016deep, zhao2016deep} since they have been shown to learn better and more general appearance features compared to hand-crafted ones \cite{krizhevsky2012imagenet}. Furthermore, networks trained on large datasets on surrogate tasks (e.g. object detection) have been shown to perform well for feature extraction on other tasks. Motivated by this, we refine VGG-16 \cite{simonyan2014very} for the detection of 10 facial AUs. More specifically, we treat the problem of AUs detection, as several binary classification problems and refine separately a VGG-16 for each AU. We replace the output layer of the VGG-16 by another with a single sigmoid unit, since each network deals with a binary classification problem. We use the binary cross-entropy as the classification cost function. That is, the total batch cost is:
\begin{equation}
C_{c}(t,q) = - \dfrac{1}{B} \sum_{b=1}^{B} (t_{b} \log q_{b} + (1-t_{b}) \log (1-q_{b})),
\end{equation}
where $B$ denotes the batch size, $t$ the target value and $q$ the predicted value.




Since the occurrence of facial expressions is correlated, many works deal with facial expression detection as a multi-label classification problem \cite{ghosh2015multi, gudi2015deep, bishay2017fusing}. In this work, we train a separate network for each expression, because the number of positive examples vary immensely from one expression to another (ranging approx. between 0.6k - 35k) -- this results in a heavily imbalanced data problem and networks that are tuned to the most populated classes. Data balancing can alleviate this problem, however, this is hard in the case of multi-label problems and typically separate networks perform better.



In total, 11 facial expressions are analysed, ten of which are facial AUs detected using the AFEA method described above, and one is smile recognized using the SmileNet proposed in \cite{jang2017smilenet}. This results in an 11-dimensional feature vector for each frame where each dimension represents the probability of one of the detected expressions.



\textbf{AFEA for the NESS Dataset.} Some patients in the NESS dataset have part of their faces occluded by wearable items e.g. have their eyes occluded by sunglasses or thick eyeglasses, or their eyebrows covered by a beanie hat. This results in wrong detection of the behaviour related to the occluded area -- typically we observe false positive activations. In order to prevent these false detections from affecting the subsequent analysis steps, for each patient/video, the mean activation over each expression is calculated and subtracted from the activations of the expression in question.




\subsection{High-Level Feature Extraction}

In section \ref{low_feat}, we extracted frame level representations, i.e. at each frame $t$ of the sequence we extracted a vector $\bm{x_{t}} \in R^{11}$,
containing the probability of the occurrence of the 11 facial expressions. In this section, we represent the set of vectors that are extracted for the whole video using a Fisher Vector (FV) representation. The FV representation is extracted by two custom DNN layers -- the first layer learns a Gaussian Mixture Model and the second layer extracts the FV representation. The first layer is first trained using a generative cost, and then both layers are refined using a discriminative cost.





We first train a \textbf{Gaussian Mixture Model (GMM)} to model the distribution of the normalized expressions probabilities $\bm{x} \in R^{11}$ using a weighted sum of $K$ Gaussian distributions \cite{sanchez2013image}. Clearly, the distribution is over the set of $\bm{x}$ that are extracted over the whole training dataset, one $\bm{x}$ for every frame of each sequence. In this context, each GMM component would represent a commonly occurring combination of facial expressions. The GMM is expressed as:
\begin{equation}
u_\lambda(\bm{x}) = \sum_{k=1}^{K}w_k u_k(\bm{x}),
\end{equation}
where $w_k$ is the weight component of the $k$-th Gaussian distribution $u_k(\bm{x})$. $u_k(\bm{x})$ is defined as:
\begin{equation}
u_k(\bm{x}) = \frac{1}{(2\pi)^{\frac{D}{2}}\lvert\bm{\Sigma}_k\rvert^{\frac{1}{2}}}\exp\left( -\frac{1}{2}(\bm{x}-\bm{\mu}_k)' \bm{\Sigma}_k^{-1} (\bm{x} - \bm{\mu}_k) \right).
\end{equation}
Each Gaussian $u_k(\bm{x})$ has three parameters associated to it, namely the weight component $w_k$, the mean vector $\bm{\mu}_k$, and the covariance matrix $\bm{\Sigma}_k$. The responsibility of each Gaussian component $u_k(\bm{x})$ in generating the input feature sample $\bm{x}_t$, is called $k$-th posterior, and is given by:
\begin{equation}
\label{eq:posterior}
\gamma_t(k)=\frac{w_k u_k(\bm{x}_t)}{\sum_{l}^{K}{w_l u_l(\bm{x}_t)}}.
\end{equation}


In this work we follow \cite{palasek2017discriminative}, and implement the GMM as a neural network layer, that during training given a set of $\bm{x}$ learns the parameters of the GMM and during testing given an $\bm{x}$ produces $K$ GMM posteriors $\lbrace \gamma_t(k), k = 1, ..., K\rbrace$ at its output (see Figure \ref{fig2}(a)). The GMM layer is first trained in unsupervised way using the Expectation-Maximization (EM) algorithm \cite{dempster1977maximum}, that is, by minimizing the negative log likelihood (i.e. the generative cost) of the complete training data.

Once the parameters of the GMM are learned, we then represent a professional-patient video interview using a \textbf{Fisher Vector (FV) representation} -- more specifically, we represent the set of low-level features, i.e. the set of vectors $\bm{x_{t}}$ extracted at each frame of the video in question, by a single high-dimensional vector (the Fisher Vector). 
The later describes how the GMM parameters should change in order to better represent the distribution of the new set of features \cite{sanchez2013image}, and is formed by stacking in a vector the gradients of the posteriors with respect to the GMM parameters; $w_k$, $\bm{\mu}_k$, and $\bm{\Sigma}_k$. Formally:
\begin{equation}
\label{eq:unnormalizedfv}
\mathscr{G}^{X}_{\lambda}=\left(\mathscr{G}^{X}_{w_1}, \ldots, \mathscr{G}^{X}_{w_K},  \mathscr{G}^{X'}_{\mu_1} \ldots,   \mathscr{G}^{X'}_{\mu_K},
\mathscr{G}^{X'}_{\sigma_1} \ldots,   \mathscr{G}^{X'}_{\sigma_K}\right)',
\end{equation}
where the gradient vectors $\mathscr{G}^{X}_{w_k}$, $\mathscr{G}^{X}_{\mu_k}$, and $\mathscr{G}^{X}_{\sigma_k}$ are calculated as follows:
\begin{equation}
\mathscr{G}^{X}_{w_k}=\left(S_{k}^{0} - Tw_k \right) / \sqrt{w_k},
\end{equation}
\begin{equation}
\mathscr{G}^{X}_{\mu_k}=\left(\bm{S}_{k}^{1} - \bm{\mu}_k S_{k}^{0} \right) / \left(\sqrt{w_k}\bm{\sigma}_k\right),
\end{equation}
\begin{equation}
\mathscr{G}^{X}_{\sigma_k}=\left(\bm{S}_{k}^{2} - 2\bm{\mu}_k \bm{S}_{k}^{1} + (\bm{\mu}_{k}^{2} - \bm{\sigma}_{k}^{2})S_{k}^{0} \right) / \left(\sqrt{2w_k}\bm{\sigma}_{k}^{2}\right),
\end{equation}
where $S_k^0$, $\bm{S}_k^1$, and $\bm{S}_k^2$ denote the 0-order, 1st-order, and 2nd-order GMM statistics, respectively, and are defined as:
\begin{equation}
\label{eq:s0}
\bm{S}_k^0=\sum_{t=1}^{T}\gamma_t(k),
\end{equation}
\begin{equation}
\label{eq:s1}
\bm{S}_k^1=\sum_{t=1}^{T}\gamma_t(k)\bm{x}_t,
\end{equation}
and
\begin{equation}
\label{eq:s2}
\bm{S}_k^2=\sum_{t=1}^{T}\gamma_t(k)\bm{x}_t^2,
\end{equation}
where $\gamma_t(k)$ is the $k$-th posterior, and $T$ is the number of local descriptors which in our case is the video length. Following \cite{sanchez2013image}, the extracted FV is normalized using both power normalization, and L2 normalization.



In \cite{palasek2017discriminative}, the FV descriptor is implemented as a neural network layer, taking as input both the GMM posteriors and VGG features, and giving as output the FV. The FV layer is used also in this work, but replacing the VGG features by the normalized probabilities of the detected expressions. The layer output or the FV has a length of $K(2N+1)$, where $K$ is the number of GMM components and $N$ is the feature dimensionality, which in our case is the number of the detected expressions, that is 11. Note that the length of the FV does not depend on the length of the video. 

Comparing the dimensionality of the low-level features (circa $500k$ for a 30-min video with 25 f/s) to the FV dimensionality ($368$ for $K=16$ and $N=11$), shows how the GMM and FV layers can efficiently reduce dimensionality. This is important in cases where the number of data samples is not very large, as is typically the case in the domain of mental illnesses.


\subsection{Regression Layers}

In order to estimate the symptom severity in schizophrenia, we use two Fully Connected (FC) layers that receive as input the output of the FV layer. The first layer ``FC1" is used for estimating individual expression-related symptoms, while the second layer ``FC2"  estimates the total scale/symptoms score (e.g. CAINS Expression scale). The number of neurons in FC1 is adjusted according to the number of the estimated symptoms in each scale (flat affect, poor rapport, and lack of spontaneity and flow of conversation symptoms in the case of PANSS and 4 Expression symptoms in the case of CAINS). Two discriminative costs are used for training the regression layers as shown in Figure \ref{fig2}(b); the first for fine-tuning the GMM with the FV and FC1 layers in an end-to-end fashion, and the second for training the FC2 layer. The mean square error is used as the discriminative cost function, and is calculated as follows:
\begin{equation}
C_{d}(p,t) = \frac{1}{V} \sum_{v=1}^{V} \frac{1}{W} \sum_{w=1}^{W} (p_{vw} - t_{vw})^2,
\end{equation}
where $V$ denotes the total number of videos/patients in our training set, $W$ is the number of symptoms estimated, and $p$ and $t$ represent the model's estimated symptom and the ground-truth value, respectively. The activation function used in FC1 and FC2 is the Rectified Linear Unit function. As the symptoms of schizophrenia have integer-based scores, the final outputs are rounded to the nearest integer during testing.





\begin{table*} [!h]
\centering
\caption{The classification results obtained by the frame-based components of the network proposed in \cite{bishay2017fusing} and by the proposed AFEA method over different facial expressions on the $15\%$ testing splits of the EmotioNet \cite{fabian2016emotionet}, ExpW \cite{SOCIALRELATION_2017}, CelebA \cite{liu2015deep}, and CEW \cite{song2014eyes} datasets.}
\label{table:101}
\begin{adjustbox}{width=0.95 \textwidth,center}
    \begin{tabular}{ | c | c | | c | c | c | c | c | c | c | c | c | c | c | c | c | c | c | c | c | c | c | c |}  \hline
  \multicolumn{2}{| c | |}{\textbf{\ \ \parbox{2cm}{ \quad \ Facial \\ Expressions}}}  & 
\textbf{\rotatebox[origin=c]{90}{\ \ \parbox{2cm}{Inner Brow \\ Raiser}}} & \textbf{\rotatebox[origin=c]{90}{\ \ \parbox{2cm}{Outer Brow \\ Raiser}}} & \textbf{\rotatebox[origin=c]{90}{\ \ \parbox{2cm}{Brow \\ Lowerer}}} & \textbf{\rotatebox[origin=c]{90}{\ \ \parbox{2cm}{Upper Lid \\ Raiser}}} & \textbf{\rotatebox[origin=c]{90}{\ \ \parbox{2cm}{Cheek \\ Raiser}}} & \textbf{\rotatebox[origin=c]{90}{\ \ \parbox{2cm}{Lip Corner \\ Puller}}} & \textbf{\rotatebox[origin=c]{90}{\ \ \parbox{2cm}{Lips Part}}}   & \textbf{\rotatebox[origin=c]{90}{\ \ \parbox{2cm}{Neutral \\ Expression}}} & \textbf{\rotatebox[origin=c]{90}{\ \ \parbox{2cm}{Lid \\ Tightener}}}  & 
\textbf{\rotatebox[origin=c]{90}{\ \ \parbox{2cm}{Eyes \\ Closed}}}  \\ \hline \hline

 \multicolumn{2}{| c | |}{\textbf{\ \ \parbox{2cm}{ \quad \ Datasets}}}  & \multicolumn{7}{| c |}{EmotioNet} & ExpW & CelebA & CEW  \\ \hline \hline

 \multirow{3}{*}{Static \cite{bishay2017fusing}}  &  Acc & 0.679  & 0.669 & 0.752 & - & 0.806 & 0.823 & 0.708 & - & 0.670 &  0.617 \\ \cline{2-12}
  
                      &  F1 & 0.166  & 0.130  & 0.354 & - & 0.544 & 0.792 & 0.697 & -  &  0.268 & 0.422 \\ \cline{2-12} 
  
					  &  Avg & 0.423  & 0.399  & 0.553 & - & 0.675 & 0.808 & 0.703 & -  &  0.469 & 0.520 \\ \hline \hline

  \multirow{3}{*}{Proposed AFEA Method} &  Acc & \textbf{0.941}  & \textbf{0.869} & \textbf{0.903} & 0.857 & \textbf{0.880} & \textbf{0.908} & \textbf{0.919} & 0.731  & \textbf{0.855} & \textbf{0.980} \\ \cline{2-12}
  
                 &  F1  & \textbf{0.459}  & \textbf{0.319}  & \textbf{0.632} & 0.304 & \textbf{0.716} & \textbf{0.897} & \textbf{0.912} & 0.718  & \textbf{0.526} & \textbf{0.977} \\ \cline{2-12} 
  
			     &  Avg & \textbf{0.700}  & \textbf{0.594}  & \textbf{0.767} & 0.580 & \textbf{0.798} & \textbf{0.902} & \textbf{0.915} & 0.725 & \textbf{0.691} & \textbf{0.979} \\ \hline 
    \end{tabular}
\end{adjustbox} 
\end{table*}


\section{Experiments and Results}

In this section, we report the performance of the developed architectures for facial expression analysis and symptom severity estimation. First, we test the performance of the proposed AFEA method and compare it with a state-of-the-art method in detecting facial expressions ``in the wild". Then, we measure the correlations between facial expressions and different symptoms of schizophrenia. Finally, we report the performance of the proposed SchiNet in estimating symptom severity and compare it to other works in the literature.


\subsection{Classification of Facial Expressions}

\label{class_AFEA}

The interviews in the NESS dataset have a wide range of different camera poses and illuminations levels. Furthermore, patients tend to gaze down or away from the interviewer, or sometimes occlude the face with different hand gestures. In order to handle with such challenges it is imperative to train the facial expression analysis method, with datasets that contain such variations -- in this work we relied on recent datasets collected ``in the wild".

\textbf{Datasets.} We use 4 datasets collected ``in the wild", for the detection of 10 facial expressions -- Table \ref{table:101} shows the used datasets, as well as the detected expressions. The facial images in these datasets were collected by searching Internet images using certain words in a variety of search engines. 
The collected images have different recording conditions and head poses -- this improves greatly the robustness of our model to those conditions. For the EmotioNet dataset \cite{fabian2016emotionet}, only manually-annotated images in the validation set are used in the training and testing of the AFEA method. The EmotioNet consists of annotations for 12 expressions. Although we trained different networks for detecting the 12 expressions, only 7 expressions (shown in Table \ref{table:101}) show good performance when applied to the NESS dataset -- those are selected for further analysis.



\textbf{Training Settings.} We split the datasets (CEW \cite{song2014eyes}, CelebA \cite{liu2015deep}, EmotioNet \cite{fabian2016emotionet}, ExpW \cite{SOCIALRELATION_2017}) into $75\%$ for training, $10\%$ for validation, and $15\%$ for testing. Many of the detected expressions have a high ratio of negative to positive examples (i.e. imbalanced data). In order to avoid the biasing of the classifier to the most frequent class (negative class), the positive and negative examples are balanced in the training set by undersampling \cite{chawla2005data}. The ExpW \cite{SOCIALRELATION_2017} dataset is annotated for 6 emotional expressions and the neutral expression. In order to keep the training set balanced and diverse when training for the detection of the neutral expression, negative examples equal to positive examples are drawn from all the 6 emotional expressions.


The training set of each expression is augmented with random flipping, rotation, shifting, shearing, and zooming, in order to avoid over-fitting. We initialize the parameters of the expression recognition networks by the parameters of the VGG-16 and refine them using SGD with adaptive learning rate (RMSprop \cite{tieleman2012lecture}), with a decay coefficient set to $0.7$ and initial learning rate to $10^{-4}$. Depending on the size of the training set for each expression, the batch size is set either to $64$ or $128$. 


\textbf{State-of-the-Art.} We compare our AFEA method with Bishay and Patras \cite{bishay2017fusing}, a method that achieves state-of-the-art results on the BP4D dataset \cite{zhang2013high} for facial expression recognition. The comparison is two-fold. First we test how both methods perform on facial expression recognition on datasets collected ``in the wild'' (Sec. \ref{class_AFEA}). Second, we apply both methods on the patients' interviews for low-level feature extraction, retrain the methods for high-level feature extraction and symptom severity estimation using the corresponding features, and observe how the performance is affected (Sec. \ref{sev_est}). In both comparisons, only the 8 expressions that are detected by both methods are used. For a fair comparison on the facial expression recognition part, we use only two of the several networks that we proposed and fused in \cite{bishay2017fusing}, and more specifically, the spatial networks that operate on the raw facial images and the coordinates of the facial landmarks without subtraction of the mean face or landmarks (i.e. CNN2, MPL2). The reason for doing so is that the network adopted here is not trained on dynamic information. In the second comparison, i.e. testing how each AFEA method performs on symptom assessment, we compare with the full architecture. 
In what follows we will refer to the simplified static version of \cite{bishay2017fusing} as ``Static \cite{bishay2017fusing}" and the full architecture as ``Full \cite{bishay2017fusing}".

\textbf{Results.} Accuracy and F1-score obtained by the  proposed method on the $15\%$ testing splits are shown in Table \ref{table:101}. We observe that the performance is highly dependent on the number of training samples and the variance in expression-appearance. More specifically, expressions like lips part, and eyes closed have a high value for both F1-score and accuracy, due to the relatively large number of training examples as well as fewer differences in expression-appearance among subjects. On other expressions like brow lowerer, and lid tightener we obtain moderately good performance due to the large variance in expression-appearance among different people. Finally, we obtain low F1-score values for the outer brow raiser and the upper lid raiser as the EmotioNet dataset is highly imbalanced for those two classes.



\begin{figure*} [t!]
\centering
\centering{\includegraphics[scale=0.49]{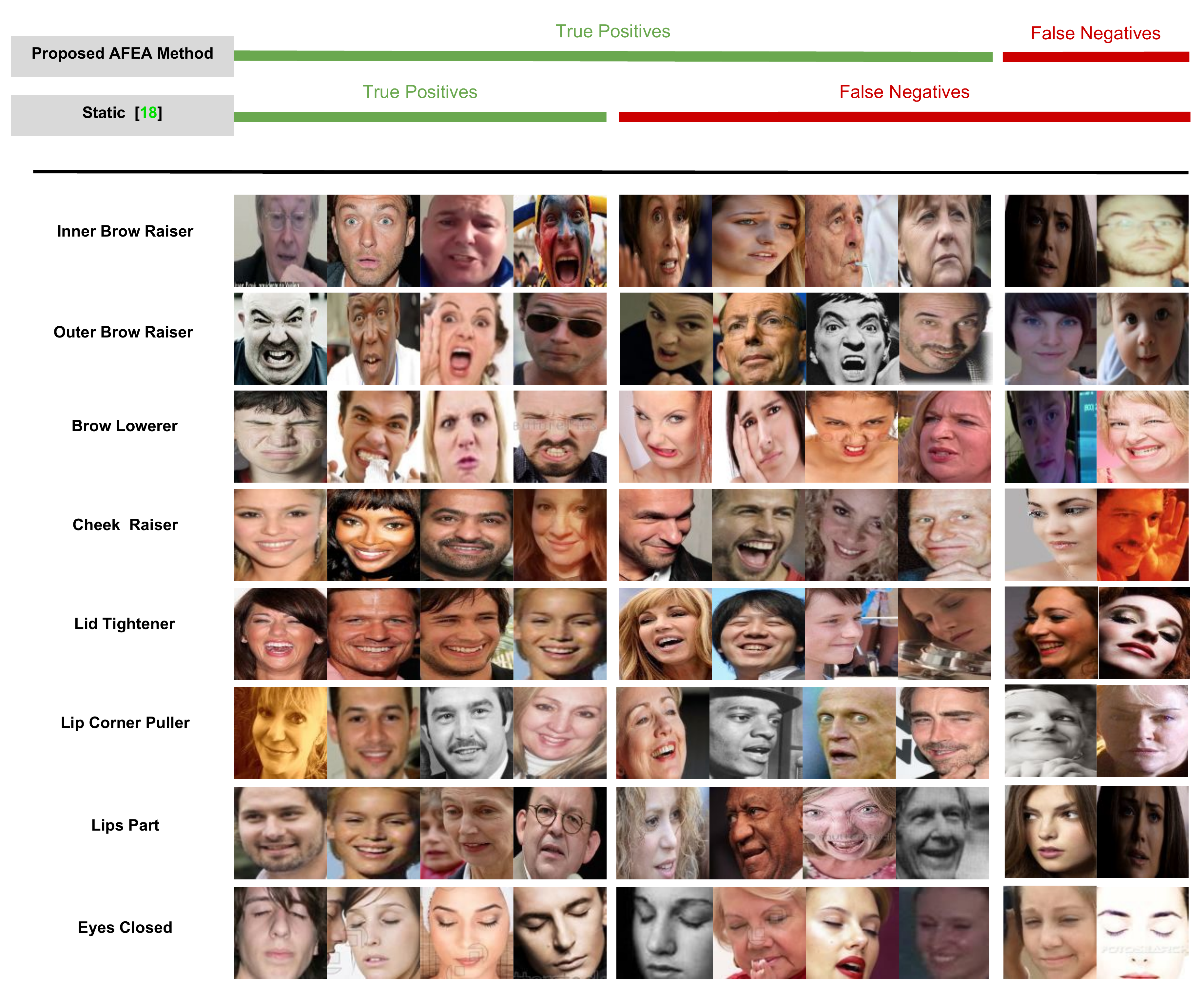}%
\label{cp1}} 
\hfil
\caption{Qualitative comparison between the proposed AFEA method, and Static \cite{bishay2017fusing} in expression analysis. Each row shows the positive examples of a certain facial expression. The true positives and false negatives achieved by each method are shown on the top part of the figure. The proposed method shows better performance in detecting expressions at several head poses and illumination conditions, compared to Static \cite{bishay2017fusing}.}
\label{fig02}
\end{figure*}

In Table \ref{table:101}, we also show the performance of Static \cite{bishay2017fusing} on the testing splits -- we observe that our method obtains better results in the 8 expressions. This considerable difference in performance is mainly due to two reasons. First, \cite{bishay2017fusing} is trained using facial images captured in a controlled environment, and with limited variation in head pose. Second, only 2 out of 8 deep networks in \cite{bishay2017fusing} are used in expressions detection. \cite{bishay2017fusing} shows that the full architecture achieves better performance than both single and combined networks. Figure \ref{fig02} shows a qualitative comparison between the two AFEA methods on different facial images drawn from the testing splits. Each image has a face with an active facial expression. We can see that Static \cite{bishay2017fusing} performs well mainly for frontal or near-frontal faces, while the proposed method can detect expressions at several head poses and illumination levels. However, the proposed method fails when the expressions are subtle, or when the faces are captured under too dark or bright illumination conditions.




\begin{table*} [!h]
\centering
\caption{Correlations found between facial expressions and the \textbf{CAINS} symptoms.}
\label{table:13}
\begin{adjustbox}{width=0.99 \textwidth,center}
\begin{threeparttable}
  \begin{tabular}{ |c| | c | c | c | c | c | c | c | c | c | c | c | c | c | c | c | c |}  \hline
  \textbf{\backslashbox[69mm]{\\ Symptoms \\ \\}{ \\ Facial \ \ \ \  \\ Expressions \\}} &
\textbf{\rotatebox[origin=c]{90}{\parbox{2cm}{Neutral \\ Expression}}} & \textbf{\rotatebox[origin=c]{90}{\parbox{2cm}{Cheek \\ Raiser}}} & \textbf{\rotatebox[origin=c]{90}{\parbox{2cm}{Lid \\ Tightener}}}  & \textbf{\rotatebox[origin=c]{90}{\parbox{2cm}{Lip Corner \\ Puller}}} & \textbf{\rotatebox[origin=c]{90}{\parbox{2cm}{Lips Part}}}   &  
\textbf{\rotatebox[origin=c]{90}{\parbox{2cm}{Smiling}}} & 
\textbf{\rotatebox[origin=c]{90}{\parbox{2cm}{Eyes \\ Closed}}}  \\ \hline \hline 
    
  \textbf{EXP -} Facial Expression & $\mathbf{0.45^{**}}$ & $\mathbf{-0.43^{**}}$ & - & $\mathbf{-0.4^{**}}$ & $-0.33^{*}$ & $\mathbf{-0.42^{**}}$ & -  \\ \hline
  
  \textbf{EXP -} Vocal Expression & $\mathbf{0.35^{**}}$ & $\mathbf{-0.38^{**}}$ & $-0.34^{*}$ &  - & $\mathbf{-0.41^{**}}$ & - & -  \\ \hline
  
  \textbf{EXP -} Expressive Gestures & - & $-0.32^{*}$ & - &  - &  $\mathbf{-0.43^{**}}$ & - & - \\ \hline
  
  \textbf{EXP -} Quantity of Speech & $\mathbf{0.38^{**}}$ & - & - & - &  $\mathbf{-0.41^{**}}$ & - & -  \\ \hline 
  
  \textbf{MAP -} Motivation for Recreational Activities & - & - & - & - &  - & - & $\mathbf{-0.47^{**}}$ \\ \hline
  
   \textbf{MAP -} Frequency of Pleasurable Recreational &  &  &  &   &  &  &    \\
   
   Activities - Past Week & - & - & - &  - & - & - & $\mathbf{-0.35^{**}}$ \\ \hline \hline
  
  \textbf{EXP -} Total Score & $\mathbf{0.42^{**}}$ & $\mathbf{-0.41^{**}}$ & - & - & $\mathbf{-0.46^{**}}$ & $-0.29^{*}$ & -  \\ \hline
    
    \end{tabular}
\begin{tablenotes}
\item $^{**}$ indicates $p \leqslant 0.001$, \quad $^{*}$ indicates $p \leqslant 0.01$
\end{tablenotes}
\end{threeparttable}

\end{adjustbox}
\end{table*}


\begin{table*} [!h]
\centering
\caption{Correlations found between facial expressions and the \textbf{PANSS} symptoms.}
\label{table:14}
\begin{adjustbox}{width=0.99 \textwidth,center}
\begin{threeparttable}
  \begin{tabular}{ |c| | c | c | c | c | c | c | c | c | c | c | c | c | c | c | c | c |}  \hline
  \textbf{\backslashbox[69mm]{\\ Symptoms \\ \\}{ \\ Facial \ \ \ \  \\ Expressions \\}} &
\textbf{\rotatebox[origin=c]{90}{\parbox{2cm}{Neutral \\ Expression}}} & \textbf{\rotatebox[origin=c]{90}{\parbox{2cm}{Inner Brow \\ Raiser}}} & \textbf{\rotatebox[origin=c]{90}{\parbox{2cm}{Outer Brow \\ Raiser}}} & \textbf{\rotatebox[origin=c]{90}{\parbox{2cm}{Cheek \\ Raiser}}} & \textbf{\rotatebox[origin=c]{90}{\parbox{2cm}{Lid \\ Tightener}}}  & \textbf{\rotatebox[origin=c]{90}{\parbox{2cm}{Lips Part}}}   & \textbf{\rotatebox[origin=c]{90}{\parbox{2cm}{Smiling}}}  \\ \hline \hline 
    
  \textbf{NEG -} Flat Affect & $0.28^{*}$ & - & - & $\mathbf{0.33^{**}}$ & - & $\mathbf{-0.37^{**}}$ & $-0.29^{*}$  \\ \hline
  
  \textbf{NEG -} Poor Rapport & - & - & - & $\mathbf{-0.36^{**}}$ & - & $-0.34^{*}$ & $-0.28^{*}$  \\ \hline
  
  \textbf{NEG -} Lack of Spontaneity and Flow of &  &  &  &   &   &  &  \\ 
  
 Conversation & $0.32^{*}$ & - & - &  - &  - & $-0.31^{*}$ & - \\ \hline 
  
  \textbf{POS -} Suspiciousness/Persecution & - & - & $\mathbf{0.36^{**}}$ & - &  - & - & -  \\ \hline 
  
  \textbf{GEN -} Somatic Concern & - & $0.29^{*}$ & - & - & $0.33^{*}$ & - & - \\ \hline
  
  \textbf{GEN -} Anxiety & - & - & $0.29^{*}$ & - & - & - & -  \\ \hline \hline
  
  \textbf{NEG -} Total Score & - & - & - & - & - & $-0.30^{*}$ & $-0.30^{*}$  \\ \hline
  
  \textbf{POS -} Total Score & - & $0.31^{*}$ & $0.30^{*}$ & - & - & $0.29^{*}$ & -  \\ \hline
  
  \textbf{GEN -} Total Score & - & - & - & - & $\mathbf{0.37^{**}}$ & - & -  \\ \hline
    
    \end{tabular}
\begin{tablenotes}
\item $^{**}$ indicates $p \leqslant 0.001$, \quad $^{*}$ indicates $p \leqslant 0.01$
\end{tablenotes}
\end{threeparttable}

\end{adjustbox}
\end{table*}

\subsection{Statistical Analysis}


The goal in this section is to calculate and examine how well a very simple feature extracted for each of the automatically detected facial expressions, namely, the frequency of the occurrence of the expression in question, correlate with the different symptom scales of schizophrenia. We show that for several symptoms, high and significant correlations with facial expressions are observed.


\textbf{Symptom Scales.} In the NESS dataset that we use, the severity of symptoms of schizophrenia is assessed by two observer-rated scales, PANSS \cite{kay1987positive}, and CAINS \cite{horan2011development}. PANSS consists of a total of 30 symptoms divided into 3 scales: Negative (NEG), Positive (POS) and General Psychopathology (GEN). Out of the 30 symptoms, 7 are grouped to form NEG scale, 7 form POS scale, and the remaining 16 symptoms form the GEN scale. Each symptom in the PANSS is rated between 1 (absent) and 7 (extreme). On the other hand, CAINS consists of 13 symptoms, divided into 2 scales: Motivation and Pleasure (MAP), and Expression (EXP). MAP has 9 symptoms and EXP has 4 symptoms. Each symptom in the CAINS has a value between 0 and 4 (0=no impairment and 4=severe impairment). 


\textbf{Calculating Correlations.} We use the Spearmans's correlation for measuring the association between the ground-truth symptom levels and activation frequency of each expression. In order to calculate the frequency, first we get a binary vector for each video frame, representing the presence or absence of each of the 11 expressions, and then compute the activation frequency as follows: 
\begin{equation}
f_{i} = \frac{N_{i}}{N_{total}}, \quad i \in \{ 1, 2, ..., 11 \}
\end{equation}
where $N_{i}$ is the number of frames for which expression $i$ is activated and $N_{total}$ is the total number of video frames with a successful face detection. 


The faces of some patients are occluded by a wearable item (e.g. thick eyeglasses) -- this sometimes results in the related facial expression being wrongly detected. In order to avoid these false detections, only frequencies that fall in the range of $-1.5 \sigma_{i} \leq f_{i} \leq 1.5 \sigma_{i}$ are considered, where $\sigma_{i}$ is the standard deviation over the frequencies of expression $i$ in the NESS dataset. Note that this step is applied only during statistical analysis and is replaced by the normalization step during symptom estimation.




\textbf{Results.} Table \ref{table:13} and \ref{table:14} show the correlations found between facial expressions on the one hand, and some symptoms in both of the CAINS and PANSS scales on the other. In CAINS (Table \ref{table:13}), significant associations are found between lips part, which is commonly activated during patients' speech, and symptoms like quantity of speech, vocal expression, and facial expression. Similarly, in PANSS (Table \ref{table:14}), higher levels of symptoms like lack of spontaneity and flow of conversation, poor rapport, and flat affect are associated with lower frequencies of the lips part. Moreover, symptoms related to the impairment in social interaction (e.g. poor rapport, flat affect, facial expression) are found to be correlated to smile and smile-related behaviour (cheek raiser). Finally, correlations are also found between facial expressions and the total score of the CAINS and PANSS scales. For instance, CAINS-EXP scale has significant associations with many facial expressions e.g. neutral expression, cheek raiser and lips part. As the NESS dataset contains patients with a relatively high level of negative symptoms and lower levels of positive and general psychopathology symptoms, more significant correlations are found with negative symptoms.



\subsection{Symptom Severity Estimation}
\label{sev_est}

Among the different types of symptoms of schizophrenia, negative symptoms are particularly difficult to assess and quantify. The assessment requires the quantification of observed verbal and especially non-verbal behaviour so that ratings commonly involve a large degree of subjectivity. Thus, an objective method for assessing these symptoms would be an important achievement. It is being debated as to what extent negative symptoms do or do not change in treatment interventions \cite{savill2015negative}, and measures that are obtained in an automatic way may establish symptoms with higher accuracy and reliability and therefore help to clarify whether changes do or do not occur. Based on that, we focus in this work on assessing the highly correlated negative symptoms in both the CAINS and PANSS interviews through automatic analysis of the video interviews.

\textbf{Training Settings.} For CAINS, the GMM and FV layers are trained firstly end-to-end with the FC1 layer for estimating the 4 EXP symptoms. Then, the GMM and FC1 parameters are fixed and the FC2 layer is trained on estimating the total EXP score. Similarly, the three highly correlated NEG symptoms in PANSS, namely flat affect, poor rapport, and lack of spontaneity and flow of conversation, are estimated at the FC1 layer, and the total NEG score is estimated at the FC2 layer. Note that the number of neurons in FC1 layer is equal to the estimated symptoms at each scale. 



The number of GMM components ($K$) is set to 16, giving for each video a FV with length 368. Following \cite{sanchez2013image}, we use variance flooring to avoid instability in the calculations -- the minimum variance allowed is 0.001. Moreover, whenever the posterior is below a threshold of $10^{-4}$ it is set to zero -- this leads to a sparser FV. The GMM-FV-FC1 layers are trained using SGD with momentum $m = 0.9$ and learning rate $lr = 0.005$ for CAINS and $0.001$ for PANSS. The FC2 layer is trained also using SGD with $m = 0.9$ and $lr = 0.01$. Finally, in the case of the CAINS scale, a scaling factor is learned for the training set and applied at testing, so as to scale the output values in the range between the minimum (0) and the maximum (4) values. Leave-One-Subject-Out (LOOCV) is used for validating and testing our architecture.





\textbf{Performance Measures.} Three measures are used for reporting the performance of the symptom severity estimation using as ground truth the psychiatrists' assessments. Following \cite{tron2015automated, tron2016facial}, we use the Pearson's Correlation Coefficient (PCC) and, in addition to it, we report the Mean Absolute Error (MAE) and the Root Mean Square Error (RMSE). The former (i.e. the MAE) is less sensitive to outliers, while the latter (i.e. the RMSE) emphasizes more on larger differences. 


\begin{table*} [!htbp]
\centering
\caption{Comparison between the proposed AFEA method, and Full \cite{bishay2017fusing} on estimating \textbf{CAINS-EXP} symptoms.}
\label{table:20}
\begin{adjustbox}{width=0.7 \textwidth,center}
    \begin{tabular}{ |c| | c | c | c | | c | c | c |}
    \hline
   & \multicolumn{3}{| c | |}{\textbf{Full \cite{bishay2017fusing}}} & \multicolumn{3}{| c |}{\textbf{Proposed AFEA Method}} \\ \cline{2-7}
 
  & PCC & MAE & RMSE & PCC & MAE & RMSE \\ \hline
 
 \textbf{EXP -} Facial Expression & 0.37 & 0.80 & 1.07 & \textbf{0.42} & \textbf{0.74} &  \textbf{0.99}  \\ \hline
 
 \textbf{EXP -} Vocal Expression & 0.25 & 0.93 & 1.22 & \textbf{0.30} & \textbf{0.75} & \textbf{1.13} \\ \hline
 
 \textbf{EXP -} Expressive Gestures & 0.04 & 1.07 & 1.37 & \textbf{0.34} & \textbf{0.99} & \textbf{1.19}  \\ \hline
 
 \textbf{EXP -} Quantity of Speech & \textbf{0.42} & 1.07 & 1.37 & 0.39 & \textbf{0.91} & \textbf{1.22}  \\ \hline \hline
 
  \textbf{EXP -} Total Score & 0.29 & 3.06 & 3.80 & \textbf{0.42} & \textbf{2.90} & \textbf{3.61}  \\ \hline

    \end{tabular}
\end{adjustbox}
\end{table*}


\begin{table*} [!htbp]
\centering
\caption{Comparison between the proposed AFEA method, and Full \cite{bishay2017fusing} on estimating \textbf{PANSS-NEG} symptoms.}
\label{table:21}
\begin{adjustbox}{width=0.7 \textwidth,center}
    \begin{tabular}{ |c| | c | c | c | | c | c | c |}
    \hline   
   & \multicolumn{3}{| c | |}{\textbf{Full \cite{bishay2017fusing} }} & \multicolumn{3}{| c |}{\textbf{Proposed AFEA Method}} \\ \cline{2-7} 
 
  & PCC & MAE & RMSE & PCC & MAE & RMSE  \\ \hline
 
 \textbf{NEG -} Flat Affect & 0.21 & 0.97 & 1.31 & \textbf{0.32} & \textbf{0.96} &  \textbf{1.27} \\ \hline
 
 \textbf{NEG -} Poor Rapport & 0.25 & 0.91 & 1.22 & \textbf{0.41} & \textbf{0.75} & \textbf{1.13} \\ \hline
 
 \textbf{NEG -} Lack of Spontaneity & & & & & & \\ 
 
 and Flow of Conversation & 0.13 & \textbf{1.28} & 1.60 & \textbf{0.24} & \textbf{1.28} & \textbf{1.54} \\ \hline  \hline
 
 \textbf{NEG -} Total Score & 0.08 & 4.00 & 4.88 & \textbf{0.40} & \textbf{3.35} & \textbf{4.27}   \\ \hline
 
    \end{tabular}
\end{adjustbox}
\end{table*}


\textbf{Effect of AFEA Architecture.} In this experiment we evaluate how our AFEA method performs on symptom severity estimation compared to \cite{bishay2017fusing}. In order to do so, the full architecture in \cite{bishay2017fusing} (Full \cite{bishay2017fusing}) is applied for analysing the videos of all of the 110 baseline patients in the NESS dataset. Out of the 110, we considered only 74 patients for whom we can successfully detect faces and landmarks in more than $90\%$ of the frames of their videos. Two output neurons are used to predict, respectively, the presence and absence of each expression in Full \cite{bishay2017fusing} -- during testing the one with the highest probability is selected. Here, in order to get an expression probability between 0-1, the presence-probability is divided by the sum of both the presence- and absence-probabilities. The mean over each expression is subtracted (normalization step), and then the expressions probabilities are used as input to the GMM layer.


As the number of patients and facial expressions analysed by the two AFEA methods vary, only the ones that are common in both methods are used in the comparison. Based on that, 69 patients and 8 common expressions (shown in Table \ref{table:101}) are selected for symptom estimation. The number of GMM components is set to 12 in this case, as fewer patients and expressions are analysed. Table \ref{table:20} and \ref{table:21} show the estimation results of the CAINS and PANSS symptoms, respectively, using both the proposed AFEA method, and the Full \cite{bishay2017fusing}. From the comparison, we can see that the proposed AFEA method leads to better symptom estimation in all the estimated symptoms. This illustrates the positive impact of training the AFEA method using data captured ``in the wild".



\begin{table*} [!htbp]
\centering
\caption{Performance of the SchiNet as well as other state-of-the-art methods on the \textbf{CAINS-EXP} symptoms.}
\label{table:15}
\begin{adjustbox}{width=0.9 \textwidth,center}
    \begin{tabular}{ |c| | c | c | c | | c | c | c || c | c | c | c |}
    \hline
   & \multicolumn{3}{| c | |}{\textbf{Tron {\em et al.} \cite{tron2015automated}}} & \multicolumn{3}{| c | |}{\textbf{Tron {\em et al.}  \cite{tron2016facial}}} & \multicolumn{3}{| c |}{\textbf{SchiNet}}  \\ \cline{2-10} 
 
  & PCC & MAE & RMSE & PCC & MAE & RMSE & PCC & MAE & RMSE \\ \hline
 
 \textbf{EXP -} Facial Expression & 0.37 & 0.80 &  1.03 & 0.36 & 0.75 &  1.07 & \textbf{0.46} & \textbf{0.66} & \textbf{0.93}  \\ \hline
 
 \textbf{EXP -} Vocal Expression & 0.23 & 0.87 & 1.23 & 0.26 & 0.86 &  1.22 & \textbf{0.27} & \textbf{0.77} & \textbf{1.10}  \\ \hline
 
 \textbf{EXP -} Expressive Gestures & 0.36 & \textbf{0.85} & 1.19 & \textbf{0.38} & 0.91 & 1.22 & 0.36 & 0.90 & \textbf{1.15}  \\ \hline
 
 \textbf{EXP -} Quantity of Speech & 0.27 & 1.09 & 1.43 & 0.25 & 1.02 &  1.36 & \textbf{0.30} & \textbf{0.98} & \textbf{1.30}  \\ \hline \hline
 

  \textbf{EXP -} Total Score & - & - & - & - & - &  - & \textbf{0.45} & \textbf{2.67} & \textbf{3.34}  \\ \hline

    \end{tabular}
\end{adjustbox}
\end{table*}


\begin{table*} [!htbp]
\centering
\caption{Performance of the SchiNet as well as other state-of-the-art methods on the \textbf{PANSS-NEG} symptoms.}
\label{table:16}
\begin{adjustbox}{width=0.9 \textwidth,center}
    \begin{tabular}{ |c| | c | c | c | | c | c | c || c | c | c | c |}
    \hline   
   & \multicolumn{3}{| c | |}{\textbf{Tron {\em et al.} \cite{tron2015automated}}} & \multicolumn{3}{| c | |}{\textbf{Tron {\em et al.} \cite{tron2016facial}}} & \multicolumn{3}{| c |}{\textbf{SchiNet}}  \\ \cline{2-10} 
 
  & PCC & MAE & RMSE & PCC & MAE & RMSE & PCC & MAE & RMSE \\ \hline
 
 \textbf{NEG -} Flat Affect & 0.37 & 0.90 &  1.28 & 0.11 & 0.99 &  1.36 & \textbf{0.42} & \textbf{0.84} & \textbf{1.18}  \\ \hline
 
 \textbf{NEG -} Poor Rapport & 0.20 & 0.98 & 1.31 & 0.15 & 1.01 &  1.26 & \textbf{0.27} & \textbf{0.85} & \textbf{1.20}  \\ \hline
 
 \textbf{NEG -} Lack of Spontaneity & & & & & & & & & \\ 
 
 and Flow of Conversation & 0.13 & 1.37 & 1.69 & 0.09 & 1.32 &  1.62 & \textbf{0.25} & \textbf{1.25} & \textbf{1.51}  \\ \hline  \hline
 
 
 \textbf{NEG -} Total Score & - & - & - & - & - & - & \textbf{0.29} & \textbf{3.30} & \textbf{4.17}  \\ \hline

    \end{tabular}
\end{adjustbox}
\end{table*}

\begin{table} [!htbp]
\centering
\caption{The estimation results of the total \textbf{PANSS-NEG} score obtained in different settings.}
\label{table:30}
\begin{adjustbox}{width=0.44 \textwidth,center}
    \begin{tabular}{ |c| | c | c | c | }
    \hline   
 \multirow{2}{*}{\textbf{Input}}  & \multicolumn{3}{| c |}{\textbf{SchiNet}}  \\ \cline{2-4} 
 
  & PCC & MAE & RMSE  \\ \hline 

 FV representation & 0.08 & 3.35 & 4.37  \\ \hline
 FC1 layer with 3 symptoms & \textbf{0.29} & \textbf{3.30} & \textbf{4.17}  \\ \hline
 FC1 layer with 7 symptoms & 0.18 & 3.37 & 4.37   \\ \hline


    \end{tabular}
\end{adjustbox}
\end{table}



\textbf{Comparison to State-of-the-Art.} In this section we compare the proposed SchiNet with two other methods that have been proposed in the literature for symptom severity estimation, namely, Tron {\em et al.} \cite{tron2015automated, tron2016facial}. We have re-implemented both methods, and for a fair comparison, the pre- and post-processing steps (e.g. normalization, scaling) applied in the SchiNet, are also applied to them. In \cite{tron2016facial}, Tron {\em et al.} used the ``Elbow criterion" for selecting the best number of clusters -- here, we tried different number of clusters in the range of 2-24, and report the best results (obtained for 12 clusters). Furthermore, since the methods in \cite{tron2015automated, tron2016facial} estimate specific symptoms of schizophrenia and not the total score, we discard the total CAINS-EXP and PANSS-NEG scores in the comparison. The 91 patients analysed by the proposed AFEA method are used in the comparison. Table \ref{table:15} and \ref{table:16} summarize the results. SchiNet outperforms the other methods in the 3 PANSS-NEG symptoms, and in 3 out of the 4 CAINS-EXP symptoms. The extracted statistical features using the GMM and FV layers show better performance compared to the hand-crafted ones.

For estimating the total PANSS-NEG score, the proposed architecture uses the 3 out of 7 PANSS-NEG symptoms that are highly correlated with facial expressions, as input to the FC2 layer. In Table \ref{table:30}, we report the estimation results in two additional settings. First,  by estimating directly the total score from the FV representation, that is by using a single fully connected layer (FC1) with a single output. Second by estimating the total score from all the PANSS-NEG symptoms. In this latter setting, we first train the SchiNet on estimating the 7 PANSS-NEG symptoms at FC1 layer, then estimating the total score at FC2 layer. In both cases, the results were worse. In the first case a possible reason is that NEG symptoms have more significant correlations with expressions than the total NEG score, so estimating symptoms first, helps a lot in estimating the total score. In the second case a possible reason is that 4 out of the 7 NEG symptoms are not correlated to facial expressions, making the training of the FC2 layer worse.


\section{Conclusions and Future Work}

Our work aims to develop an automatic tool that is capable of quantifying patient behaviour, and then using it for estimating the severity of different symptoms. To this end, interviews of symptom assessment recorded at different places in the UK were used in our analysis, in conditions that are similar to real clinical settings. Analysing interviews of patients at a wide variety of poses and illumination conditions led us to implement an AFEA method that is trained using data collected ``in the wild", that is outside laboratory conditions. Then, patients' facial expressions are detected and used as input to a neural network, that extracts compact statistical features and estimates symptoms of schizophrenia. We estimate expression-related negative symptoms in two different assessment interviews, PANSS and CAINS.


Our experimental results show many findings. First, we show that the proposed method for AFEA ``in the wild" performs better on symptom estimation than another state-of-the-art method \cite{bishay2017fusing} that was trained using data captured in a controlled environment. This underlines the importance of training with data collected ``in the wild". Second, significant correlations are found between symptoms and the frequency of occurrence of automatically detected facial expressions -- this confirms that symptom levels of patients with schizophrenia are expressed in the degree of their impairments in expression of emotion and social interaction. Third, several symptoms in the PANSS and CAINS interviews can be estimated with a MAE less than 1 level. All of that leads to a conclusion that quantified patient behaviour with a well-trained deep architecture is a feasible and reliable method for estimating negative symptoms of schizophrenia -- the latter is a challenging task in clinical settings -- and may be used as an objective method to establish changes during treatment.


Although our architecture shows promising results in symptom estimation, comparing the correlations between the automatic estimations and professional assessment (reaching at most to $0.46$), to the correlations between assessments of different professionals that have annotated the NESS dataset \cite{priebe2016effectiveness} (equals to $0.85$), shows that automatic estimation of symptom severity needs further improvement to reach human level performance. In order to improve the performance of symptom severity estimation, we plan in our future work to improve the performance of the AFEA method, by moving from static to temporal analysis ``in the wild". Moreover, we will extend the behaviour analysis to include body gestures and vocal expressions (besides facial expressions). Finally, we will explore how the symptom estimation can be improved using a personalized model.

\ifCLASSOPTIONcompsoc
  \section*{Acknowledgments}
\else
  \section*{Acknowledgment}
\fi

The work of Mina Bishay is a part of the Newton-Mosharafa PhD scholarship, which is jointly funded by the Egyptian Ministry of Higher Education and the British Council.


\ifCLASSOPTIONcaptionsoff
  \newpage
\fi



\bibliographystyle{IEEEtran}
\bibliography{egbib}
%




%

\begin{IEEEbiography}[{\includegraphics[width=1in,height=1.25in,clip,keepaspectratio]{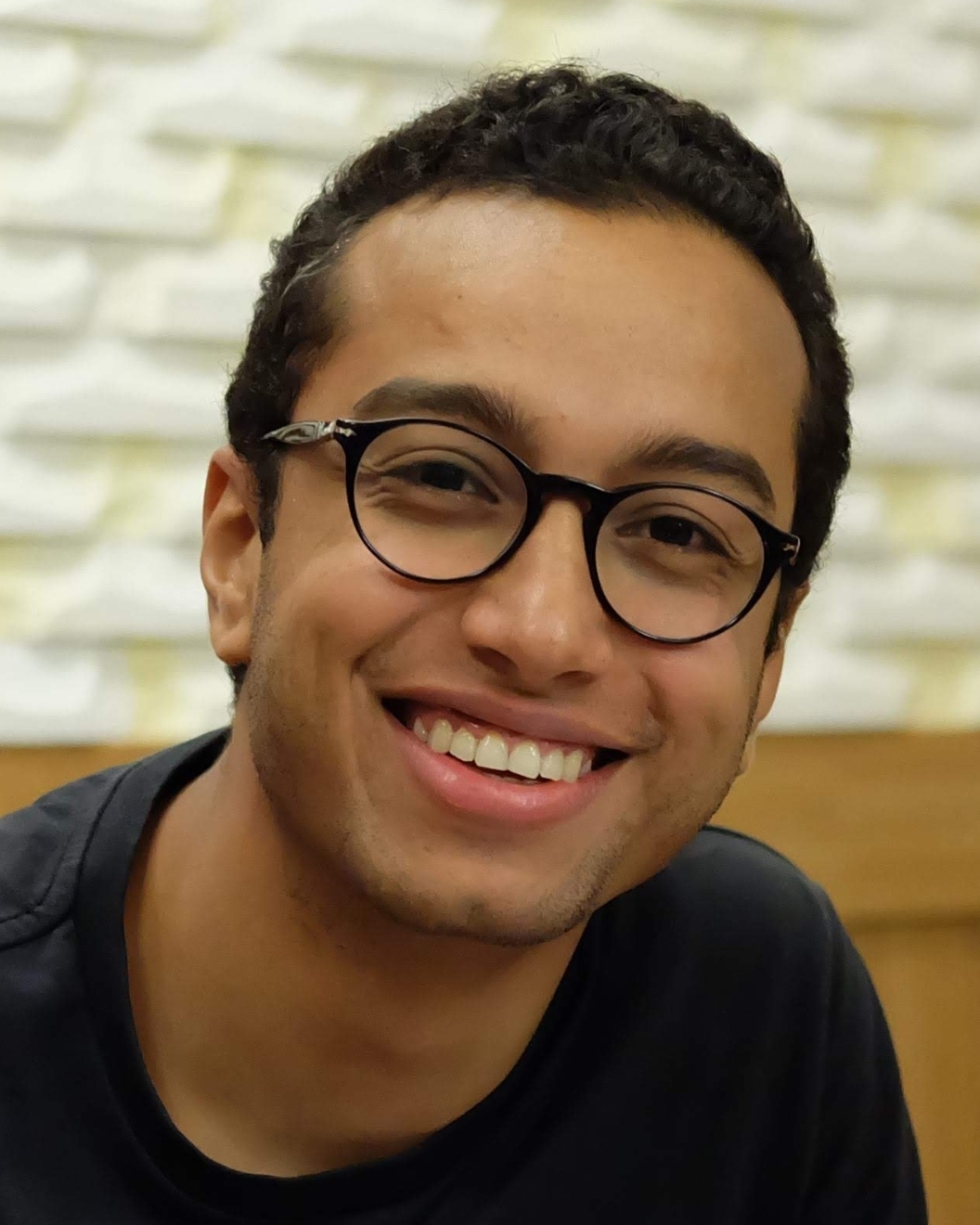}}]{Mina Bishay}
received the BSc and MSc degrees (with honors) in electrical engineering (Electronics and Communications section) from Assiut University, Egypt, in 2010 and 2014, respectively. During his MSc thesis, he focused on improving the performance of fingerprint image segmentation and enhancement. He received the Best Student Paper Award in IEEE National Radio Science Conference (NRSC) in 2014.  He is now working towards the PhD degree at the School of Electronic Engineering and Computer Science, Queen Mary University of London, UK. His research interests include: automatic behaviour analysis, affective computing and deep learning. 
\end{IEEEbiography}

\begin{IEEEbiography}[{\includegraphics[width=1in,height=1.25in,clip,keepaspectratio]{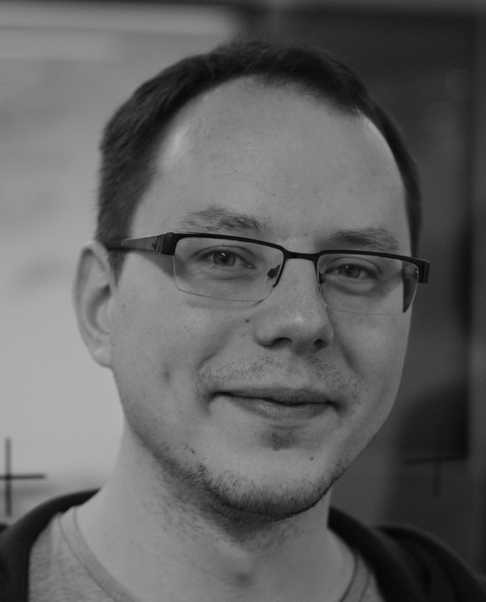}}]{Petar Palasek}
received the BSc and MSc degrees in computer science from the Faculty of Electrical Engineering and Computing, University of Zagreb in 2010 and 2012 respectively, and the PhD degree from the School of Electronic Engineering and Computer Science, Queen Mary University of London in 2017. During his PhD, his research focused on studying deep learning architectures for human action recognition in videos, and it also included the problems of 3D human reconstruction and human body pose estimation. His research interests include computer vision, machine learning and neural networks. He is currently working as a research scientist at MindVisionLabs LTD, a London based startup.
\end{IEEEbiography}

\begin{IEEEbiography}[{\includegraphics[width=1in,height=1.25in,clip,keepaspectratio]{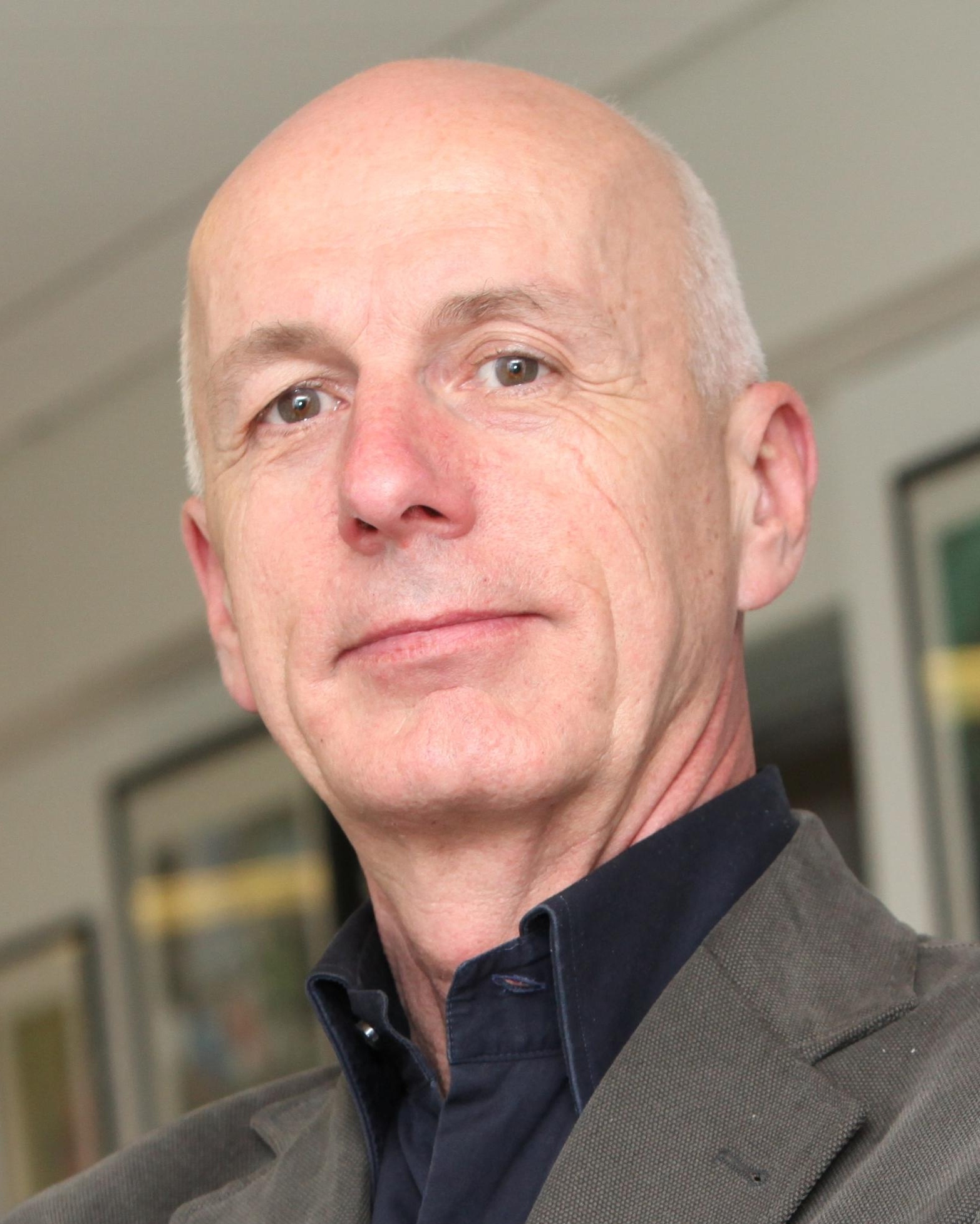}}]{Stefan Priebe} 
graduated in Psychology and Medicine, and qualified as Neurologist, Psychiatrist and Psychotherapist in Germany. Since 1997 he has been Professor for Social and Community Psychiatry at Queen Mary, University of London (QMUL). He is also Director of the WHO Collaborating Centre for Mental Health Service Development, Director of the NIHR Global Health Research Group for Developing Psycho-Social Interventions, Deputy Director of a registered Clinical Trials Unit, Research Director of the Institute for Population Health Sciences (all at QMUL) and R\&D Director of East London NHS Foundation Trust. 

He heads a research group in East London which runs several programmes focusing on understanding, modifying and utilising social interactions to reduce mental distress. 
\end{IEEEbiography}

\begin{IEEEbiography}[{\includegraphics[width=1in,height=1.25in,clip,keepaspectratio]{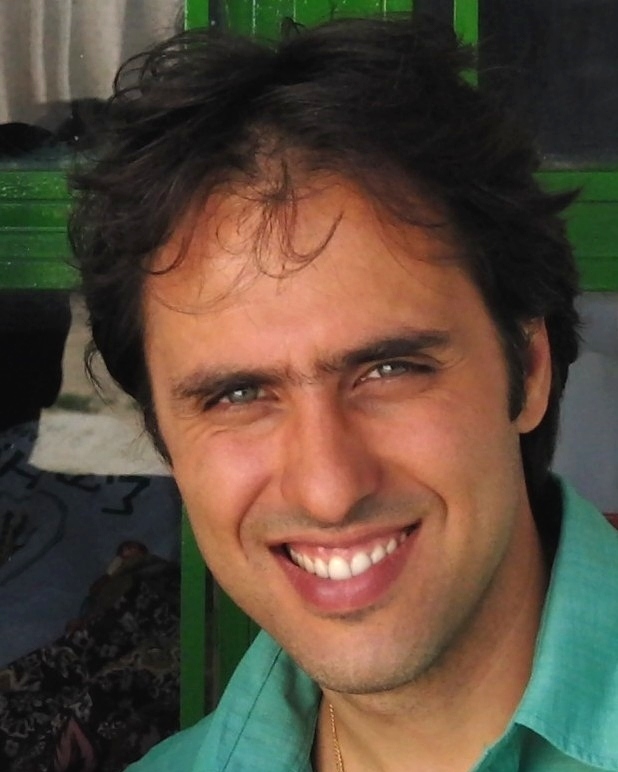}}]{Ioannis Patras}
received the BSc and MSc degrees in computer science from the Computer Science Department, University of Crete, Heraklion, Greece, in 1994 and 1997, respectively, and the PhD degree from the Department of Electrical Engineering, Delft University of Technology, Delft (TU Delft), The Netherlands, in 2001. He is a Reader in the School of Electronic Engineering and Computer Science
Queen Mary University of London, London, UK. His current research interests are in the area of Computer Vision, Machine Learning and Affective Computing, with emphasis on the analysis of visual data depicting humans and their activities. He is an Associate Editor of the Image and Vision Computing Journal, Pattern Recognition, and Computer Vision and Image Understanding.
\end{IEEEbiography}







\end{document}